%% file: root.tex
\documentclass[letterpaper, 10 pt, conference]{ieeeconf}  

\IEEEoverridecommandlockouts                              

\usepackage{amsmath,amsthm,amssymb,bm}
\usepackage{graphicx,fixltx2e}
\usepackage{hyperref}
\usepackage{xy}
\usepackage[]{algorithm2e}
\usepackage{bbm}
\usepackage{todonotes}
\usepackage{float}
\usepackage{cite}
\usepackage{cleveref}

\newcommand{\N}{\mathbb{N}}
\newcommand{\E}{\mathbb{E}}

\newcommand{\R}{\mathbb{R}}

\makeatletter
\newcommand{\pushright}[1]{\ifmeasuring@#1\else\omit\hfill$\displaystyle#1$\fi\ignorespaces}
\newcommand{\pushleft}[1]{\ifmeasuring@#1\else\omit$\displaystyle#1$\hfill\fi\ignorespaces}
\usepackage{hyperref}
\makeatletter
\def\@citex[#1]#2{\leavevmode
\let\@citea\@empty
\@cite{\@for\@citeb:=#2\do
{\@citea\def\@citea{,\penalty\@m\ }%
\edef\@citeb{\expandafter\@firstofone\@citeb\@empty}%
\if@filesw\immediate\write\@auxout{\string\citation{\@citeb}}\fi
\@ifundefined{b@\@citeb}{\hbox{\reset@font\bfseries ?}%
\G@refundefinedtrue
\@latex@warning
{Citation `\@citeb' on page \thepage \space undefined}}%
{\@cite@ofmt{\csname b@\@citeb\endcsname}}}}{#1}}
\makeatother

\newenvironment{lemma}[2][Lemma]{\begin{trivlist}
\item[\hskip \labelsep {\bfseries #1}\hskip \labelsep {\bfseries #2.}]}{\end{trivlist}}

\newenvironment{problem}[2][Problem]{\begin{trivlist}
\item[\hskip \labelsep {\bfseries #1}\hskip \labelsep {\bfseries #2.}]}{\end{trivlist}}

\allowdisplaybreaks

\title{\LARGE \bf
Non-Gaussian Risk Bounded Trajectory Optimization for Stochastic Nonlinear Systems in Uncertain Environments
}

\author{Weiqiao Han*, Ashkan Jasour*, Brian Williams
\thanks{Computer Science and Artificial Intelligence Laboratory, Massachusetts Institute of Technology, \{weiqiaoh,jasour,williams\}@mit.edu.
This work was partially supported by the Boeing grant 6943358. *These
authors contributed equally to the paper.}%
}

\begin{document}

\newcounter{eqn}

\maketitle
\thispagestyle{empty}
\pagestyle{empty}

\begin{abstract}
\input{sec/abstract.tex}
\end{abstract}

\section{Introduction}
\input{sec/introduction.tex}

\section{Problem Definition}
\input{sec/problem_definition}

\section{Method}
\input{sec/method}

\section{Experiments}
\input{sec/experiments}

\section{Conclusion}
\input{sec/conclusion}

\newpage

\bibliographystyle{IEEEtran}
\bibliography{references}

\end{document}

%% file: sec/abstract.tex
We address the risk bounded trajectory optimization problem of stochastic nonlinear robotic systems. More precisely, we consider the motion planning problem in which the robot has stochastic nonlinear dynamics and uncertain initial locations, and the environment contains multiple dynamic uncertain obstacles with arbitrary probabilistic distributions. The goal is to plan a sequence of control inputs for the robot to navigate to the target 
while bounding the probability of colliding with obstacles. Existing approaches to address risk bounded trajectory optimization problems are limited to particular classes of models and uncertainties such as Gaussian linear problems. 
In this paper, we deal with stochastic nonlinear models, nonlinear safety constraints, and arbitrary probabilistic uncertainties, the most general setting ever considered. 
To address the risk bounded trajectory optimization problem, we first formulate the problem as an optimization problem with stochastic dynamics equations and chance constraints. 
We then convert probabilistic constraints and stochastic dynamics constraints on random variables into a set of deterministic constraints on the moments of state probability distributions. Finally, we solve the resulting deterministic optimization problem using nonlinear optimization solvers and get a sequence of control inputs. 
To our best knowledge, it is the first time that the motion planning problem to such a general extent is considered and solved. 
To illustrate the performance of the proposed method, we provide several robotics examples.



%% file: sec/introduction.tex
In order to bring robots into everyday life, it is critical to plan trajectories for robots to navigate safely in uncertain environments. 
However, the motion planning problem in dynamic environments is computationally hard even in its simplest form \cite{reif1994motion}.
In this paper, we consider the motion planning problem to its most generality, where the environment contains multiple dynamic uncertain obstacles with arbitrary probabilistic distributions and the robot itself has nonlinear stochastic dynamics and uncertain initial locations. Our goal is to plan a sequence of control inputs for the robot to navigate to the target, while bounding the probability of colliding with obstacles.

Standard motion planning algorithms, such as rapidly exploring random tree (RRT), probabilistic roadmap (PRM), and virtual potential field methods, plan safe trajectories under environments with deterministic obstacles \cite{ramanujam1987planning, lynch2017modern}. 
Planning algorithms under uncertainties usually plan trajectories that have bounded probability of collision with obstacles.  
Existing planning algorithms under probabilistic uncertainties are mainly limited to Gaussian uncertainty and convex obstacles \cite{axelrod2018provably, schwarting2017parallel, dai2019chance, lew2020chance, dawson2020provably, luders2010chance, blackmore2009convex} or rely on sampling to estimate the probabilistic safety constraints \cite{blackmore2010probabilistic, janson2018monte, calafiore2006scenario, cannon2017chance}.
For example, \cite{axelrod2018provably} proposes a variant of the RRT planning algorithm that plans a provably safe path for robots under obstacles with Gaussian uncertainty without considering the dynamical constraints. Also, \cite{lew2020chance, dawson2020provably, luders2010chance} use convex obstacles, while \cite{blackmore2009convex} assumes the feasible state space is a convex polytope.
Some authors \cite{schwarting2017parallel, dai2019chance, lew2020chance, dawson2020provably, luders2010chance, blackmore2009convex} consider system dynamics in planning under uncertainty. For example, \cite{schwarting2017parallel} assumes deterministic system dynamics, while \cite{dai2019chance, lew2020chance, dawson2020provably, luders2010chance, blackmore2009convex} consider linearized system dynamics with Gaussian uncertainty. 

Methods that do not make Gaussian assumptions are usually based on sampling. 
For example, \cite{blackmore2010probabilistic} approximates the distribution of the system state using a finite number of particles.
\cite{janson2018monte} proposes a variance-reduced Monte Carlo probability estimation algorithm for path collision probability computation.
\cite{calafiore2006scenario, cannon2017chance} use the scenario approach in which they sample the constraints to obtain a standard convex optimization problem (the scenario problem) whose solution is approximately feasible for the original set of constraints.
Sampling approaches can be computationally inefficient and do not guarantee to satisfy the probabilistic constraints. 

Recently, moment-based approaches, such as \cite{jasour2021convex,wang2020non,jasour2021real, wang2020fast,jasour2018moment,jasour2019risk}, use higher order statistics of probability distributions to plan safe trajectories and verify probabilistic constraints without making the Gaussian uncertainty and/or convex obstacle assumptions. For example, \cite{jasour2021convex} uses analytic inequalities to bound the risk of collision, converting the chance constraints to constraints on moments of probabilities. It then uses standard motion planning algorithms, such as RRT, aided by sum-of-squares (SOS) verification to plan safe trajectories using convex optimization. However, \cite{jasour2021convex} does not consider system dynamics. \cite{wang2020non} uses similar concentration inequalities to relax the chance constraints and solves relaxed nonconvex trajectory optimization using nonlinear solvers, but it considers convex obstacles and does not assume uncertainties in system dynamics or initial positions. 

When considering stochastic dynamical systems, in order to describe the uncertainty of system states over the planning horizon, the problem of uncertainty propagation arises. 
For linear systems with Gaussian noise, one can use the prediction step of the Kalman filter for uncertainty propagation \cite{kalman1960new}. 
Similarly, the extended Kalman filter\cite{sorenson1985kalman} and the unscented Kalman filter \cite{julier1997new} generalize uncertainty propagation of the Kalman filter to nonlinear systems with Gaussian noise by considering linearized models and samples of uncertainties (sigma points), respectively. When dealing with nonlinear systems with non-Gaussian noise, Monte Carlo-based methods, such as particle filter \cite{del1997nonlinear, del1998measure},
are commonly used. These methods use a set of particles to represent and propagate uncertainties.
However, they are computationally expensive and non-deterministic.

Recently, \cite{jasour2021moment} addresses the exact uncertainty propagation problem for nonlinear systems with non-Gaussian noise based on moments of the probability distributions. The method is able to compute polynomial, trigonometric, and mixed-trigonometric-polynomial moments up to any desired order over the planning horizon. We will apply this method to our optimization, converting stochastic dynamics constraints into deterministic constraints on moments.

In this paper, we consider the motion planning problem to its most generality. We 
consider the following setting: (1) The system dynamics is nonlinear and stochastic, and the uncertainty is not necessarily Gaussian; (2) The initial position of the system is uncertain and does not necessarily have Gaussian distribution; (3) The obstacle can be of arbitrary shape, can deform over time, can move, and has arbitrary uncertainty. 
We propose a novel approach to address the risk bounded trajectory optimization problem in this general setting.
We first formulate the planning problem as an optimization problem with stochastic dynamics constraints and chance constraints. We then convert the optimization problem into a standard nonlinear optimization problem by converting chance constraints and stochastic dynamics constraints into deterministic constraints on moments. We solve the resulting nonconvex optimization in the variable of moments using nonlinear optimization solvers to obtain a sequence of control inputs that steer the system to the target region. 
To our best knowledge, it is the first time that the motion planning problem to such a general extent is considered and solved. 
We demonstrate our method on several robotics examples. 

%% file: sec/problem_definition.tex
Let $\mathcal{X}\subseteq \R^n$ be the state space and $\mathcal{U} \subseteq \R^m$ be the control input space, where $n, m \in \N^+$.
The system dynamics is 
\begin{equation}
    \mathbf{x}_{t+1} = f(\mathbf{x}_t, \mathbf{u}_t, \mathbf{\omega}_t)\label{eq:sysdyn}
\end{equation}
where $\mathbf{x} \in \mathcal{X}, \mathbf{u} \in \mathcal{U}$ are the state and control input at the current time step, respectively, and $\mathbf{\omega}$ is the disturbance with known probability distribution, not necessarily Gaussian. 
The initial state $\mathbf{x}_0$ is a random variable with some known probability distribution. 
The obstacles are denoted by $\mathcal{O}_i, i = 1,\ldots, M$, where $M\in \N^+$, and each obstacle $\mathcal{O}_i$ can be represented by polynomials 
\begin{align}\label{eq:obs}
    \mathcal{O}_i(\tilde{w}_i, t) = \{\mathbf{x} \in \mathcal{X} : \  p_i(\mathbf{x}, \tilde{\omega}_i, t) \leq 0  \}, \ i=1,...,M
\end{align}
where $p_i$ is a polynomial and $\tilde{\omega}_i$ is a random variable with some known probability distribution. 
The time variable $t$ in the polynomial $p_i$ indicates that the obstacle can change its position and shape over time.
We define the risk to be the probability of collision with any uncertain obstacles at any time step, and the probability of not reaching the goal region $\mathbf{x}_{goal}\subseteq \mathcal{X}$ at the final time step. 
The goal of risk bounded trajectory optimization is to find a sequence of control inputs $u_0, \ldots, u_{T-1}$ over the time horizon $T \in \N^+$ 
to minimize the cost function and bound the risk.
More precisely, the risk bounded trajectory optimization problem can be formulated as the following probabilistic optimization:
\begin{problem}{1} Risk Bounded Trajectory Optimization
\begin{equation}
\begin{aligned}
\min_{\mathbf{u}} \quad &\E[l_f(\mathbf{x}_T) + \sum_{t=0}^{T-1} l(\mathbf{x}_t, \mathbf{u}_t,\omega_t)]\\
\text{s.t.} \quad & \mathbf{x}_{t+1} = f(\mathbf{x}_t, \mathbf{u}_t, \omega_t), \  \ |_{t = 0}^{T-1},\\
& \text{Prob}(\mathbf{x}_t \in \mathcal{O}_i(\tilde{\omega}_i,t)) \leq \Delta, \ \  |_{t = 0}^{T-1}, \ |_{i=1}^{M}\\
& \text{Prob}(\mathbf{x}_T \not\in \mathbf{x}_{goal}) \leq \Delta_{goal},\\
& \mathbf{x}_0 \sim pr(\mathbf{x}_0)
\end{aligned}
\end{equation}
where $\text{Prob}(\mathbf{x}_t \in \mathcal{O}_i(\tilde{\omega}_i,t))$ and $\text{Prob}(\mathbf{x}_T \not\in \mathbf{x}_{goal})$ are risks, $\Delta,\Delta_{goal} \in [0,1]$ are the given acceptable risk levels, and $pr(\mathbf{x}_0)$ is the given probability distribution of the initial system states. 
\end{problem}
In this paper, we will use moments of probability distributions to represent uncertainties and convert the probabilistic trajectory optimization problem into a deterministic optimization problem.

\subsection{Notations and Definitions}
Let $\R[x]$ be the ring of polynomials in the variables $\mathbf{x} = (x_1,\ldots,x_n)$ with real coefficients. 
A polynomial $p(\mathbf{x}) \in \R[x]$ can be written as $p(\mathbf{x}) = \sum_{\bm{\alpha}\in\N^n} p_{\bm{\alpha}} \mathbf{x}^{\bm{\alpha}}$, where $\bm{\alpha} = (\alpha_1,\ldots,\alpha_n) \in \N^n$ and $\mathbf{x}^{\bm{\alpha}} = \prod_{i=1}^n x_i^{\alpha_i}$ is a monomial in standard basis. 

Let $(\Omega, \Sigma, \mu)$ be a probability space, where $\Omega$ is the sample space, $\Sigma$ is the $\sigma$-algebra of $\Omega$, and $\mu: \Sigma \to [0,1]$ is the probability measure on $\Sigma$. 
Suppose $\mathbf{x} \in \Omega \subseteq \R^n$ is an $n$-dimensional random vector. 
Let $\bm{\alpha} = (\alpha_1,\ldots,\alpha_n)\in \N^n$.
The expectation of $\mathbf{x}^{\bm{\alpha}}$ defined as $m_{\bm{\alpha}} = \mathbb{E}[\mathbf{x}^{\bm{\alpha}}] $ is a {moment of order $\alpha$}, where $\alpha = \sum_i \alpha_i$.
The sequence of all moments of order $\alpha$, denoted by $\mathbf{m}_\alpha$, is the expectation of all monomials of order $\alpha$ sorted in graded reverse lexicographic order (grevlex).
For example, the sequence of moments of order $\alpha = 3$ of random vector $\mathbf{x} \in \mathbb{R}^2$ is 
$\mathbf{m}_3 = [m_{3,0}, m_{2,1}, m_{1,2}, m_{0,3}] = [\mathbb{E}[x_1^3], \mathbb{E}[x_1^2x_2], \mathbb{E}[x_1x_2^2], \mathbb{E}[x_2^3]]$. 

We will use characteristic functions and trigonometric functions to describe the moments of nonlinear functions of the motion dynamics \cite{jasour2021moment}. 
For any random variable $\mathbf{x}$ with a probability density function, the characteristic function always exists and is defined as
\begin{align}
    \Phi_\mathbf{x}(t) = \mathbb{E}[e^{it^\top \mathbf{x}}]
\end{align}
Trigonometric polynomials of order $n$ are defined as $p(x) = \sum_{k=0}^n a_k \cos(kx) + b_k \sin(kx)$, where $a_k, b_k \in \R, k=0,\ldots,n$ are the constants.
Mixed trigonometric polynomials are defined as $p(x) = \sum_{k=0}^n a_k x^{b_k} \cos^{c_k}(x) \sin^{d_k}(x)$, where $a_k \in \R$ and $b_k, c_k, d_k \in \N, k=0,\ldots,n$ are constants.

%% file: sec/method.tex
Our method consists of three steps. 
First, we replace the chance constraints with a set of deterministic constraints in terms of the moments of the state probability distributions. Any state probability distribution whose moments satisfy the obtained deterministic constraints is guaranteed to satisfy the chance constraints of the planning problem.
The second step is the uncertainty propagation where we replace the stochastic dynamics equation constraints by moment propagation equation constraints. This allows us to describe the moments of the state probability distributions at each time step in terms of the control inputs. The first two steps turn the chance constrained optimization problem into a nonlinear deterministic optimization problem. Finally, we solve the resulting nonlinear optimization problem using the off-the-shelf nonlinear optimization solvers.

\subsection{Chance Constraints}
In this section, we use the notion of risk contours to transform  the  probabilistic safety constraints of the planning  problem  in to a set of deterministic constraints in terms of the moments of the state probability distributions. In \cite{Contour1,jasour2021convex,jasour2021real}, we define the risk contour as the set of all states whose  probability  of collision with  the  uncertain  obstacle  is bounded by $\Delta$. In this paper, given that states of the autonomous system are also uncertain, we define the moment-based risk contour as the set of all moments whose probability distribution satisfies the probabilistic safety constraint. More precisely, we define the moment-based risk contour at time $t$ with respect to the moments of the uncertain state and the uncertain obstacle as follows:
\begin{equation}\label{rc_s_1}
\mathcal{M}^{\Delta}_{r_i}(t):= \{\ \mathbf{m}_{\alpha}(t)|_{\alpha=0}^{2d}: \ \hbox{Prob}( \mathbf{x}_{t} \in \mathcal{O}_i(\tilde{\omega},t)) \leq \Delta  \} \ |_{i=1}^{M}
\end{equation}
where, $\mathbf{m}_{\alpha}(t)|_{\alpha=0}^{2d}$ are the moments of order $\alpha=0,...,2d$ of the uncertain state $\mathbf{x}_t$.
To construct the moment-based risk contour, we replace the probabilistic constraint, i.e.,
$\hbox{Prob}( \mathbf{x}_{t} \in \mathcal{O}_i(\tilde{\omega},t)) =\hbox{Prob}( p_i(\mathbf{x}_{t}, \tilde{w}_i, t)\leq 0 ) \leq \Delta$, with a deterministic constraint in terms of the moments of state $\mathbf{x}_t$. In this paper, we provide an analytical method as follows:

Given the polynomial $p_i(\mathbf{x}, \tilde{w}_i, t) $ of the uncertain obstacle $\mathcal{O}_i(\tilde{\omega},t) $, we define the set $\hat{\mathcal{M}}_{r_i}^{\Delta}(t)$, $i=1,...,M$ as follows:
\begin{align}\label{rc_s_2}
&\hat{\mathcal{M}}^{\Delta}_{r_i}(t):= \notag \\
&\left\lbrace \ 
\mathbf{m}_{\alpha}(t)|_{\alpha=0}^{2d}: 
\begin{array}{cc} 
\frac{4}{9}\frac{\mathbb{E}[p_i^2(\mathbf{x}, \tilde{\omega}_i, t)] - \mathbb{E}[p_i(\mathbf{x}, \tilde{\omega}_i, t)]^2}{\mathbb{E}[p_i^2(\mathbf{x}, \tilde{\omega}_i, t)]} \leq \Delta, \\ 
\mathbb{E}[p_i(\mathbf{x}, \tilde{\omega}_i, t)]^2\geq \frac{5}{8} \mathbb{E}[p_i^2(\mathbf{x}, \tilde{\omega}_i, t)]  , \\
\mathbb{E}[p_i(\mathbf{x}, \tilde{\omega}_i, t)]\geq 0   
\end{array}
\right\rbrace 
\end{align}
where the expectation is taken with respect to the distribution of uncertain states and uncertain parameter $\tilde{\omega}_i$. Note that we can compute $\mathbb{E}[p_i^2(\mathbf{x}, \tilde{\omega}_i, t)]$ and $\mathbb{E}[p_i(\mathbf{x}, \tilde{\omega}_i, t)]$ in \eqref{rc_s_2} in terms of the moments of uncertain state $\mathbf{x}_t$ and known moments of $\tilde{\omega}_i$.
The following result holds true.

\textbf{Theorem 1:} The set $\hat{\mathcal{M}}_{r_i}^{\Delta}$ in \eqref{rc_s_2} is an inner approximation of the moment-based risk contour ${\mathcal{M}}_{r_i}^{\Delta}$ in \eqref{rc_s_1}.

\textit{Proof}: 
To obtain constraints in \eqref{rc_s_2}, we use concentration inequalities that provide bounds on how a random variable deviates from a certain value. Concentration inequalities can be used to relax the chance constraints in the original chance constrained optimization problem \cite{jasour2021convex, wang2020non}.
More precisely, we will leverage one-sided  Vysochanskij–Petunin  (VP) inequality defined for unimodal random variable $z \in \mathbb{R}$ as follows \cite{mercadier2021one}: 
\begin{align}
    P(z - \mathbb{E}[z] \geq r) \leq \begin{cases}
    {4 \over 9} {\sigma^2 \over \sigma^2 + r^2} & \text{for } r^2 \geq {5\over 3}\sigma^2\\
    {4 \over 3} {\sigma^2 \over \sigma^2 + r^2} - {1\over 3} & \text{otherwise}
    \end{cases}
\end{align}
where $\sigma$ is the variance and $r \geq 0$. We now define random variable $z \in \mathbb{R}$ in terms of the polynomial of the uncertain obstacle as $z= -p_i(\mathbf{x},\tilde{\omega}_i,t)$ and $r = \mathbb{E}[p_i(\mathbf{x},\tilde{\omega}_i,t)]$. By applying the VP inequality, we obtain the constraints in \eqref{rc_s_2}. 
Given that the probability provided by VP inequality is an upper bound of the probability of the safety constraint, the set in \eqref{rc_s_2} is an inner approximation of the moments of probability distributions that satisfy the safety probabilistic safety constraints. 
Also, the obtained upper bound of the chance constraint is closely related to upper bound approximation of the indicator function of the safety constraint (For more information see \cite{jasour2021convex}).\hfill $\blacksquare$




\textbf{Example 1}: The set {$\mathcal{O}= \left\lbrace  (x_1,x_2) :  x_1^2+x_2^2-\tilde{\omega}^2 \leq 0 \right\rbrace $} 
represents a circle-shaped obstacle whose radius $\tilde{\omega}$ has a uniform probability distribution over {$[0.3,0.4]$}, Moment of order $\alpha$ of a uniform distribution defined over {$[l,u]$} can be described in a closed-form as {$\frac{u^{\alpha+1}-l^{\alpha+1}}{(u-l)(\alpha+1)}$}. To construct the moment-based risk contour in \eqref{rc_s_2}, we can compute {$\mathbb{E}[p^2(\mathbf{x}, \tilde{\omega}, t)]$} and {$\mathbb{E}[p(\mathbf{x}, \tilde{\omega}, t)]$} in terms of the moments of uncertain states $\mathbf{x}$ and known moments of $\tilde{\omega}$ as follows:
\begin{align}
    \mathbb{E}[p] &= \mathbb{E}[x_1^2+x_2^2-\tilde{\omega}^2 ] =-0.123+m_{2,0}+m_{0,2} \notag\\
     \mathbb{E}[p^2] &= \mathbb{E}[\left(x_1^2+x_2^2-\tilde{\omega}^2 \right)^2] = 0.015 - 0.246m_{2,0} \notag \\ 
    &\quad\quad  - 0.246m_{0,2} + m_{4,0} + 2m_{2,2} + m_{0,4} \notag
\end{align}
where $m_{i,j}=\mathbb{E}[x_1^ix_2^j]$ is the moment of order $(i+j)$ of the uncertain state $\mathbf{x}$.

\begin{figure}[t]
    \centering
    \includegraphics[width=0.5\textwidth]{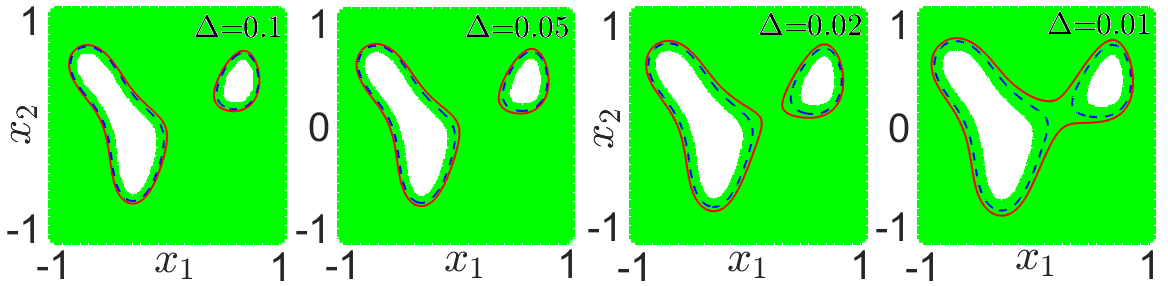}
    \caption{ Sets of all states $(x_1,x_2)$ with bounded risk of $\Delta$: True set (green), inner approximation obtained using VP based risk contour in \eqref{rc_s_2} (outside of the dashed-curve ), and inner approximation obtained by the method in \cite{jasour2021convex} (outside of the solid-curve)}
    \label{fig:RC1}
\end{figure}

\textbf{Example 2}:
An uncertain obstacle is described by the polynomial $p(\mathbf{x},\tilde{\omega})=- 0.42x_1^5 - 1.18x_1^4x_2 - 0.47x_1^4 + 0.3x_1^3x_2^2 - 0.57x_1^3x_2 + 0.6x_1^3 - 0.65x_1^2x_2^3 + 0.17x_1^2x_2^2 + 1.87x_1^2x_2 + 0.06x_1^2 + 0.69x_1x_2^4 - 0.14x_1x_2^3 - 0.85x_1x_2^2 + 0.6x_1x_2 - 0.21x_1 + 0.01x_2^5 - 0.06x_2^4 - 0.07x_2^3 - 0.41x_2^2 - 0.08x_2 +0.07 - 0.1w $ where the uncertain parameter $\omega$ has a Beta distribution with parameters $(9,0.5)$ over $[0,1]$. 
In this example, we assume that states $(x_1,x_2)$ are deterministic and obtain the set of all states whose probability of collision with the uncertain obstacle in bounded by $\Delta$. Figure \ref{fig:RC1} compares the VP-inequality-based risk contour of this paper with the provided risk contour in \cite{jasour2021convex} for different risk levels $\Delta$. As shown in Figure \ref{fig:RC1}, VP based risk contour results in less conservative safe sets.

\subsection{Uncertainty Propagation}
In \cite{jasour2021moment}, we provide an exact moment-based uncertainty propagation method through robotic systems. In this section, we leverage \cite{jasour2021moment} to describe the moments of uncertain states in terms of the control inputs.

Given the stochastic dynamical system \eqref{eq:sysdyn}, we are going to compute the moments of order $\alpha \in \N$ of the state $\mathbf{x}_t$ over the time horizon $t = 0, 1,\ldots, T$.
We assume that $f$ only contains certain elementary functions, including polynomials, trigonometric functions, and mixed-trigonometric-polynomial functions. 
This assumption is not conservative, given that many robotic dynamical systems can be represented by these elementary functions. For example, these classes of functions can capture the translational motions of the robot represented by the linear terms as well as rotational motions of the robot represented by the trigonometric terms of the rotation matrix.
To be able to compute the moments of the uncertain states, we need to compute the moments of the nonlinear terms of function $f$ in \eqref{eq:sysdyn}. For this purpose, we will use the following lemmas to compute the moments of trigonometric and mixed-trigonometric-polynomial functions.

\begin{lemma}{1}\cite[Lemma 2]{jasour2021moment}
Let $\theta$ be a random variable with characteristic function $\Phi_\theta(t)$. Given $(\alpha_1, \alpha_2) \in \N^2$ where $\alpha = \sum_{i=1}^2 \alpha_i$, trigonometric moments of order $\alpha$ of the form $m_{c_{\theta}^{\alpha_1}s_{\theta}^{\alpha_2}} = \mathbb{E}[\cos^{\alpha_1}(\theta) \sin^{\alpha_2}(\theta)]$ can be computed as 
\begin{align}
    {(-i)^{\alpha_2} \over 2^{\alpha_1+\alpha_2}}
    &\sum_{(k_1,k_2) = (0,0)}^{(\alpha_1,\alpha_2)}
    {\alpha_1\choose k_1}{\alpha_2\choose k_2}\times \nonumber \\
    &(-1)^{\alpha_2-k_2} \Phi_{\theta}(2(k_1+k_2) - \alpha_1 - \alpha_2)
\end{align}
\end{lemma}

\begin{lemma}{2} \cite[Lemma 4]{jasour2021moment}
Let $\theta$ be a random variable with characteristic function $\Phi_\theta(t)$. Given $(\alpha_1, \alpha_2, \alpha_3) \in \N^3$ where $\sum_{i=1}^3 \alpha_i = \alpha$, mixed-trigonometric-polynomial moments of order $\alpha$ of the form 
$m_{\theta^{\alpha_1} c_{\theta}^{\alpha_2} s_{\theta}^{\alpha_3}} = \mathbb{E}[\theta^{\alpha_1} \cos^{\alpha_2}({\theta}) \sin^{\alpha_3}({\theta})]$ can be computed as
\begin{align}
    {1\over i^{\alpha_1 + \alpha_3} 2^{\alpha_2 + \alpha_3}}
    &\sum_{(k_1,k_2) = (0,0)}^{(\alpha_2,\alpha_3)}
    {\alpha_2\choose k_1}{\alpha_3\choose k_2}\times \nonumber\\
    &(-1)^{\alpha_3 - k_2}{d^{\alpha_1} \over dt^{\alpha_1}} \Phi_\theta(t) |_{t=2(k_1+k_2) -\alpha_2 -\alpha_3}
\end{align}
\end{lemma}

To compute the moments of the uncertain states, we will construct a new deterministic dynamical system that governs the exact time evolution of the moments of the uncertain states. For this purpose, we will first 
construct the augmented system defined as
\begin{align} 
    \mathbf{x}_{aug_{t+1}} = A(\mathbf{u}, \omega, t) \mathbf{x}_{aug_{t}} 
\end{align}
where $\mathbf{x}_{aug} $ is a vector of mixed-trigonometric-polynomial basis generated by $\mathbf{x}_t$ and $A(\mathbf{u}, \omega, t) $ is a matrix of nonlinear functions of $\mathbf{u}$ and $\omega$.
Now, using the definition of standard moments of order $\alpha$, we can recursively describe the moments of order $\alpha$ of the uncertain states at time $t+1$, i.e., $\mathbb{E}[\mathbf{x}^{\alpha}_{aug_{t+1}} ]$, in terms of the moments at time $t$, i.e., $\mathbb{E}[\mathbf{x}^{\alpha}_{aug_{t}}]$. More precisely,  we get the moment propagation equation as follows
\begin{align}
    \mathbf{m}_{\alpha}(t+1) = A_{mom_{\alpha}}(t) \mathbf{m}_{\alpha}(t) \label{eq:mom_prop}
\end{align}
where $\mathbf{m}_{\alpha}$ is the vector of moments of the augmented state $\mathbf{x}_{aug}$ and $A_{mom_{\alpha}}(t)$ is the matrix of nonlinear functions of $\mathbf{u}$ and known moments of $\omega$.

\textbf{Example 3:} Consider the following stochastic nonlinear system:
\begin{align} 
\begin{split}
x_{t+1}&=x_t+v_t\cos\left(\theta_t\right) \\
\theta_{t+1}&=\theta_t+\omega_{{\theta}_t}
\notag
\end{split}
\end{align}
\noindent where, $(x,\theta)$ are the states, $v$ is the control input, and $\omega_{\theta}$ is the external disturbance. We define the augmented state vector as $\mathbf{x}_{aug}=[x,\cos(\theta),\sin(\theta)]^T$. By doing so, the augmented system for the nonlinear stochastic system is obtained as follows: 
\begin{equation}\label{exa5_aug}
\mathbf{x}_{aug_{t+1}}=
\begin{bmatrix}
1 & v_t & 0\\
0 & \cos(\omega_{{\theta}_t}) & -\sin(\omega_{{\theta}_t})\\
0 & \sin(\omega_{{\theta}_t}) & \cos(\omega_{{\theta}_t})
\end{bmatrix}\mathbf{x}_{aug_t} 
\end{equation} 
The obtained augmented system is equivalent to the original nonlinear stochastic system. 
Using the definition of the moments and the augmented system, we obtain the moment systems of the form \eqref{eq:mom_prop} that govern the exact time evolution of the moments of the uncertain states of the nonlinear stochastic system. For example, we obtain the moment system of order $\alpha=1$ of the form $ \mathbf{m}_{1}(t+1)=A_{mom_{1}}(t)\mathbf{m}_{1}(t)$
for the augmented system in \eqref{exa5_aug} as follows:
\begin{equation}
\mathbf{m}_{1}(t+1)=
\begin{bmatrix}
1 & v_t & 0\\
0 & m_{c_{\omega_{\theta}}}(t) & -m_{s_{\omega_{\theta}}}(t)\\
0 & m_{s_{\omega_{\theta}}}(t) & m_{c_{\omega_{\theta}}}(t)
\end{bmatrix}\mathbf{m}_{1}(t)
\notag 
\end{equation} 
where, $\mathbf{m}_{1}(t)=\left[ \ \mathbb{E}[x_t],\  \mathbb{E}[\cos(\theta_t)],\  \mathbb{E}[\sin(\theta_t)]  \  \right ]^T$ is the vector of all moments of order $\alpha=1$ of $\mathbf{x}_{aug_t}$. Also,  matrix $A_{mom_{1}}(t)$ is described in terms of the control input and known first order trigonometric moments of the external disturbance.

Similarly, we obtain the moment system of order $\alpha=2$ of the form $\mathbf{m}_{{2}}(t+1)=A_{mom_{2}}(t)\mathbf{m}_{{2}}(t)
$ where $\mathbf{m}_{2}(t)$ is the vector of all moments of order
$\alpha=2$ of $\mathbf{x}_{aug}$ as follows:
{$\mathbf{m}_{2}(t) = \left[ \
\mathbb{E}[x^2_t], \ 
\mathbb{E}[x_t\cos(\theta_t)], \ 
\mathbb{E}[x_t\sin(\theta_t)],
\mathbb{E}[\cos^2(\theta_t)],\right.$}
{
$\left. \mathbb{E}[\cos(\theta_t)\sin(\theta_t)], \
\mathbb{E}[\sin^2(\theta_t)] \   \right]^T \notag
$}. Also, matrix $A_{mom_{2}}$ is obtained in terms of the control input and known first and second order trigonometric moments of the external uncertainty as follows:
\begin{center}
    \resizebox{1\linewidth}{!}{%
$A_{mom_{2}}(t)=\begin{bmatrix}
 1&     2v_t&         0&              v^2_t&                       0&                  0\\
 0& m_{c_{\omega_\theta}}(t)& -m_{s_{\omega_\theta}}(t)&       v_tm_{c_{\omega_\theta}}(t)&            -v_tm_{s_{\omega_\theta}}(t)&                  0\\
 0& m_{s_{\omega_\theta}}(t)&  m_{c_{\omega_\theta}}(t)&       v_tm_{s_{\omega_\theta}}(t)&             v_tm_{c_{\omega_\theta}}(t)&                  0\\
 0&        0&         0&        m_{c^2_{\omega_\theta}}(t)&    -2m_{c_{\omega_\theta}s_{\omega_\theta}}(t)&         m_{s^2_{\omega_\theta}}(t)\\
 0&        0&         0& m_{c_{\omega_\theta}s_{\omega_\theta}}(t)& m_{c^2_{\omega_\theta}}(t) - m_{s^2_{\omega_\theta}}(t)& -m_{c_{\omega_\theta}s_{\omega_\theta}}(t)\\
 0&        0&         0&        m_{s^2_{\omega_\theta}}(t)&     2m_{c_{\omega_\theta}s_{\omega_\theta}}(t)&         m_{c^2_{\omega_\theta}}(t)\\
 \end{bmatrix} 
$ }
\end{center}

\subsection{Final Optimization Problem}
To obtain the deterministic optimization of the planning problem, we replace the stochastic dynamics equation by the moment propagation equation \eqref{eq:mom_prop}.
We also replace the constraints on risk of collisions by deterministic constraints in terms of the moments in \eqref{rc_s_2}. Similarly, we can replace the constraint on probability of reaching the goal region with deterministic constraints in terms of the moments, i.e., $\mathcal{M}_{r_g}^{\Delta_{goal}}(T)$. 

By doing so, we arrive at an deterministic nonlinear optimization problem where the decision variables are control inputs $\mathbf{u}$ and moments of the augmented states as follows:
\begin{problem}{2} Deterministic Trajectory Optimization
\begin{equation}
\begin{aligned}
\min_{\mathbf{m_{\alpha}}|_{\alpha=1}^{2d},\mathbf{u}} \quad &\E[l_f(\mathbf{x}_T) + \sum_{t=0}^{T-1} l(\mathbf{x}_t, \mathbf{u}_t,\omega_t)]\\
\text{s.t.} \quad & \mathbf{m}_{\alpha}(t+1) = A_{mom_{\alpha}}(t) \mathbf{m}_{\alpha}(t),\ |_{t = 0}^{T-1},\\
& \mathbf{m}_{\alpha}(t) \in \mathcal{M}_{r_i}^{\Delta}(t), \  |_{t = 0}^{T-1}, \ |_{i = 1}^{M},\\
& \mathbf{m}_{\alpha}(T) \in \mathcal{M}_{r_g}^{\Delta_{goal}}(T),\\
& \mathbf{m}_{\alpha}(0) = \mathbb{E}[\mathbf{x}^{\alpha}_0]
\end{aligned}
\end{equation}
\end{problem}
We can solve the obtained nonlinear optimization using the off-the-shelf interior point solvers. 
The size of the obtained deterministic optimization is a function of 1) the number of the planning time steps, 2) the highest order of the polynomials of obstacles and the goal region, and 3) the number of system states.


%% file: sec/experiments.tex
In this section, we demonstrate our method on several robotics examples. 
The computations in this section were performed on a computer with Intel i7 2.6 GHz processors and 16 GB RAM.
We use Casadi package \cite{andersson2019casadi} for Matlab as the wrapper to solve nonlinear optimization problems. The optimization returns a sequence of controls and a sequence of moments of the uncertain system states over the planning horizon\footnote{The code is on  \url{github.com/jasour/Non-Gaussian_Risk-Bounded_TrajOpt}}. In the experiments, we plot the expected value of the obtained state trajectory and also expected obstacle locations using polynomial $p_i(\mathbf{x}, \mathbb{E}[\tilde{\omega}_i], t)$ of \eqref{eq:obs}.




\subsection{Underwater Vehicle}\label{exp1}
\begin{figure}
    \centering
    \includegraphics[width=0.3\textwidth]{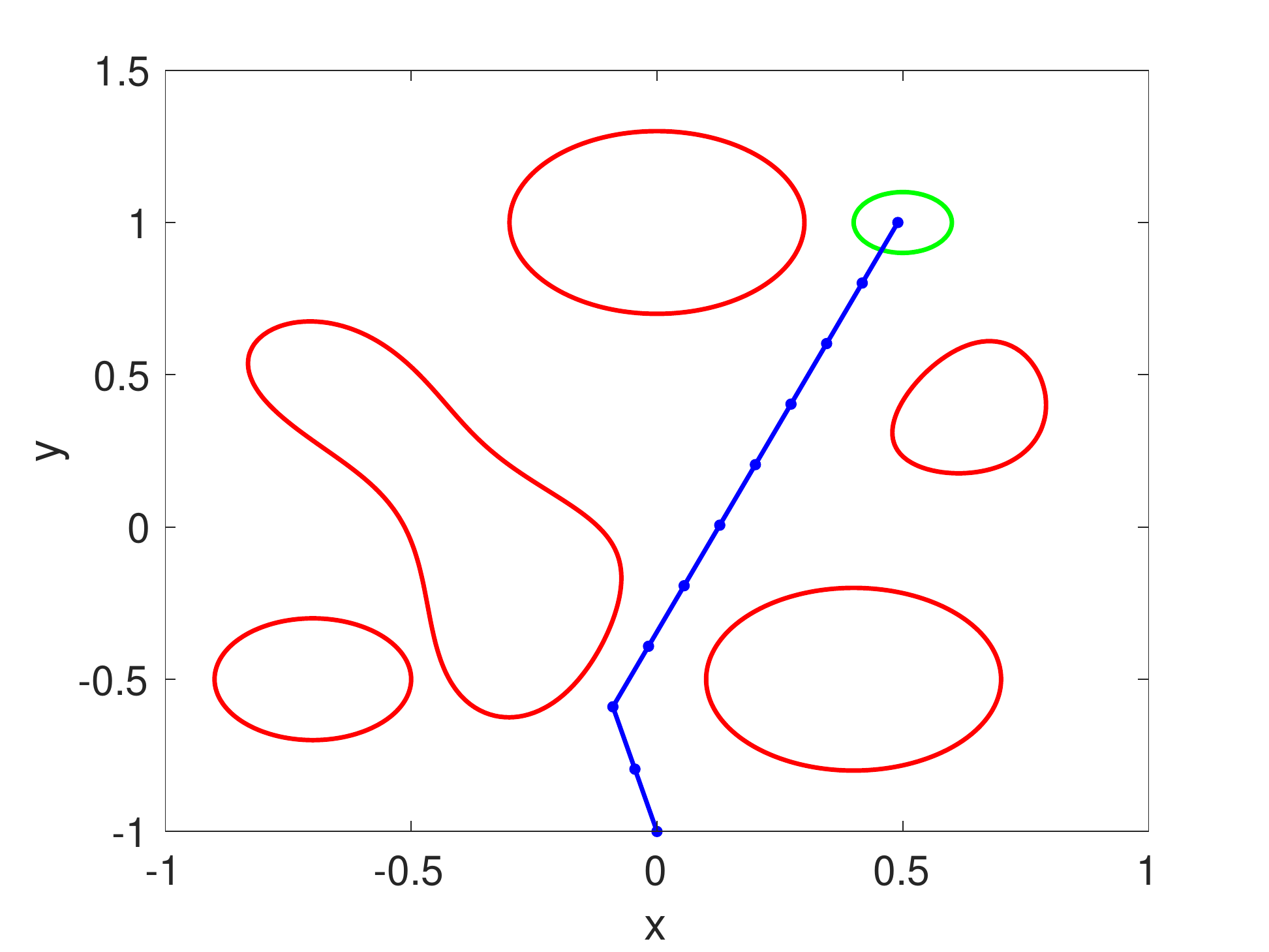}
    \caption{Example \ref{exp1}: The red curves are the expected location of the obstacles, where the high order obstacle consists of two regions. The green curve defines the goal region. The blue path is the expected optimal path. The trajectory has the bounded risk of 0.1, and the system enters the goal region with probability of at least 0.99. }
    \label{fig:exp1}
\end{figure}
Motion of an underwater vehicle in the presence of external disturbances is modeled as
\begin{align*}
    &x_{t+1} = x_{t} + \Delta T (v_t + {\omega_v}_t) \cos(\theta_t + {\omega_{\theta}}_t)\\
    &y_{t+1} = y_t + \Delta T (v_t + {\omega_v}_t) \sin(\theta_t + {\omega_{\theta}}_t)
\end{align*}
where $(x,y)$ is the position, and $\Delta T = 0.1$. We model the external disturbances using the random variables $\omega_v$ and $\omega_\theta$ with uniform distribution on $[-0.1, 0.1]$. We can control the position of the vehicle using its velocity $v$ and steering angle $\theta$. Given the uncertain initial location of the vehicle modeled by uniform distribution on $[-0.1, 0.1]\times[-0.1, 0.1]$, we want to guide the vehicle toward the goal region in $T=10$ steps while minimizing the objective function $\sum_{t = 0}^T v_t^2 + \theta_t^2$
and avoiding the uncertain unsafe regions. The acceptable risk bounds of the planning problem are $\Delta=0.1, \Delta_{goal}=0.1$.

The goal region is defined by the polynomial inequality of the form of \eqref{eq:obs} with polynomial $p_g(\mathbf{x}) = (x_1 - 0.5)^2 + (x_2 - 1)^2 - 0.1^2 \leq 0$. Uncertain unsafe regions are modeled by 4 polynomial inequalities in the form of \eqref{eq:obs} with the following polynomials: 
$p_1(\mathbf{x},\omega_1)= 0.42x_1^5 + 1.18x_1^4x_2 + 0.47x_1^4 - 0.3x_1^3x_2^2 + 0.57x_1^3x_2 - 0.6x_1^3 + 0.65x_1^2x_2^3 - 0.17x_1^2x_2^2 - 1.87x_1^2x_2 - 0.06x_1^2 - 0.69x_1x_2^4 + 0.14x_1x_2^3 + 0.85x_1x_2^2 - 0.6x_1x_2 + 0.21x_1 - 0.01x_2^5 + 0.06x_2^4 + 0.07x_2^3 + 0.41x_2^2 + 0.08x_2 -0.07 + 0.1\omega_1 \leq 0$, $p_2(\mathbf{x},\omega_2) = (x_1-0.4)^2+(x_2+0.5+\omega_2)^2 - 0.3^2 \leq 0$, 
$p_3(\mathbf{x},\omega_3) = x_1^2+(x_2+1.0+\omega_3)^2 - 0.3^2 \leq 0$, and
$p_4(\mathbf{x},\omega_4) = (x_1+0.7)^2+(x_2+0.5+\omega_4)^2 - 0.2^2 \leq 0$, where $\omega_1$ has Beta distribution with parameters $(9,0.5)$ over $[0,1]$, 
and $\omega_2,\omega_3,\omega_4$ have uniform distribution on $[-0.02, 0.02]$.

We obtain the moment propagation equations up to order $2d$, where $d=5$ is the maximum polynomial order of the unsafe regions.
The optimization has around 700 variables and around 900 constraints.
The optimization variables include 2D state moments up to order 10 and two control variables for each time step. 
It took the solver around 158 seconds to find the optimal solution.
\Cref{fig:exp1} shows the planned trajectory. 
To verify the results, we estimate the risk of the obtained trajectory using Monte Carlo simulation. We sample one million points from the initial distribution and simulate them forward via the stochastic dynamics.
The trajectory is verified to have the bounded risk of 0.1, and the system enters the goal region with probability of at least 0.99.


\subsection{Aerial Vehicle}\label{exp2}
\begin{figure}[t]
    \centering
    \includegraphics[width=0.2\textwidth]{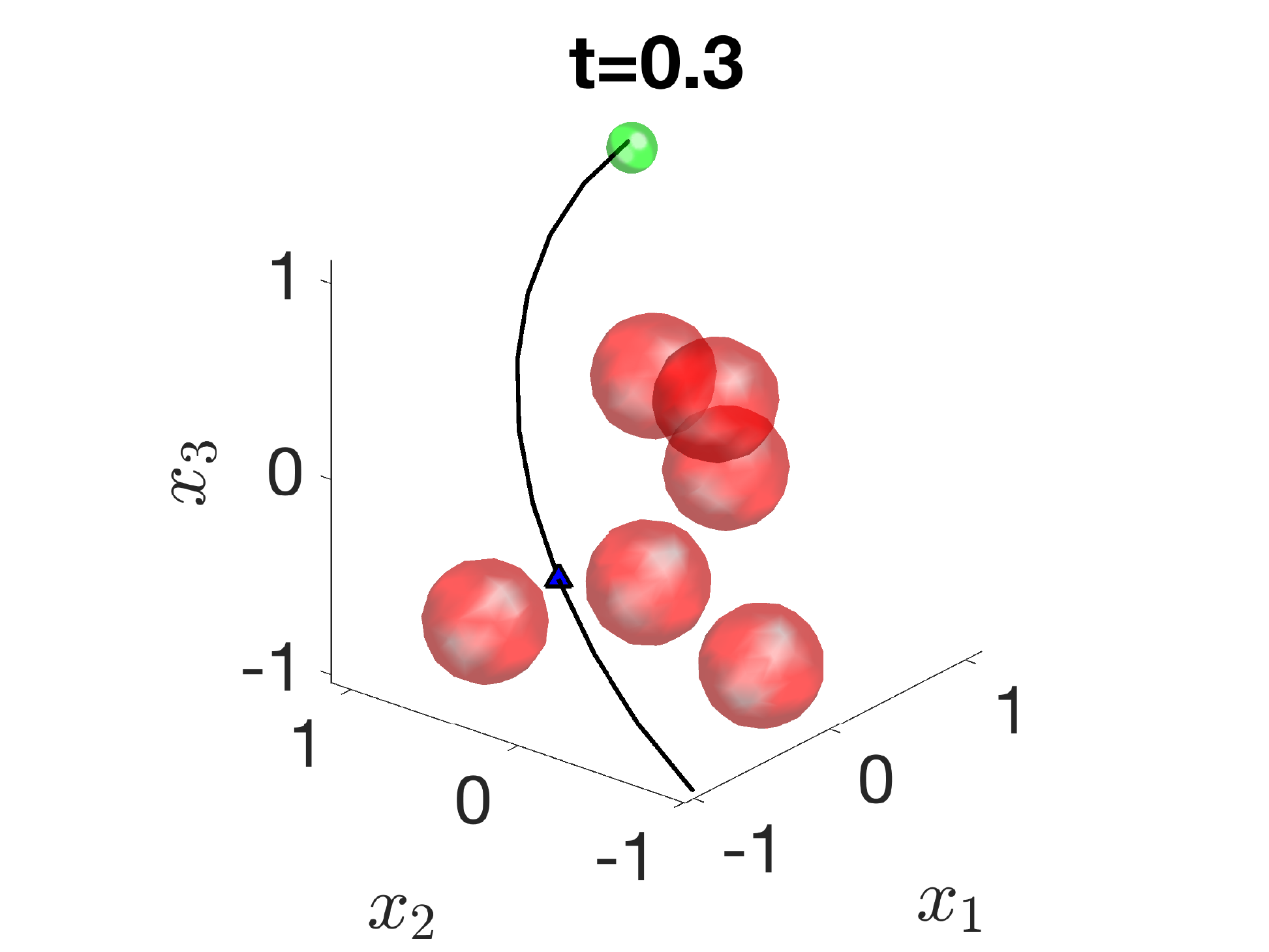}
    \includegraphics[width=0.2\textwidth]{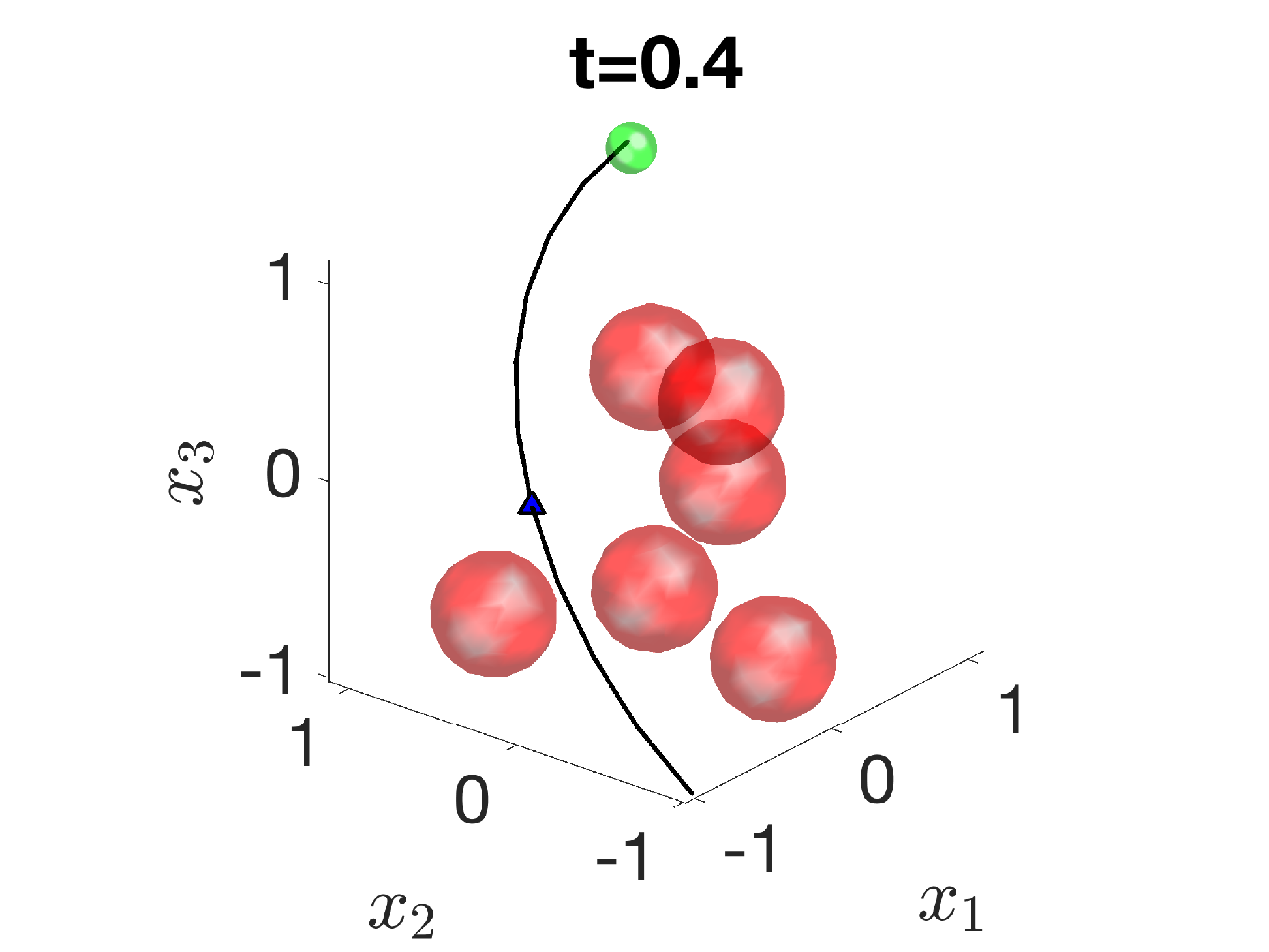}
    \includegraphics[width=0.2\textwidth]{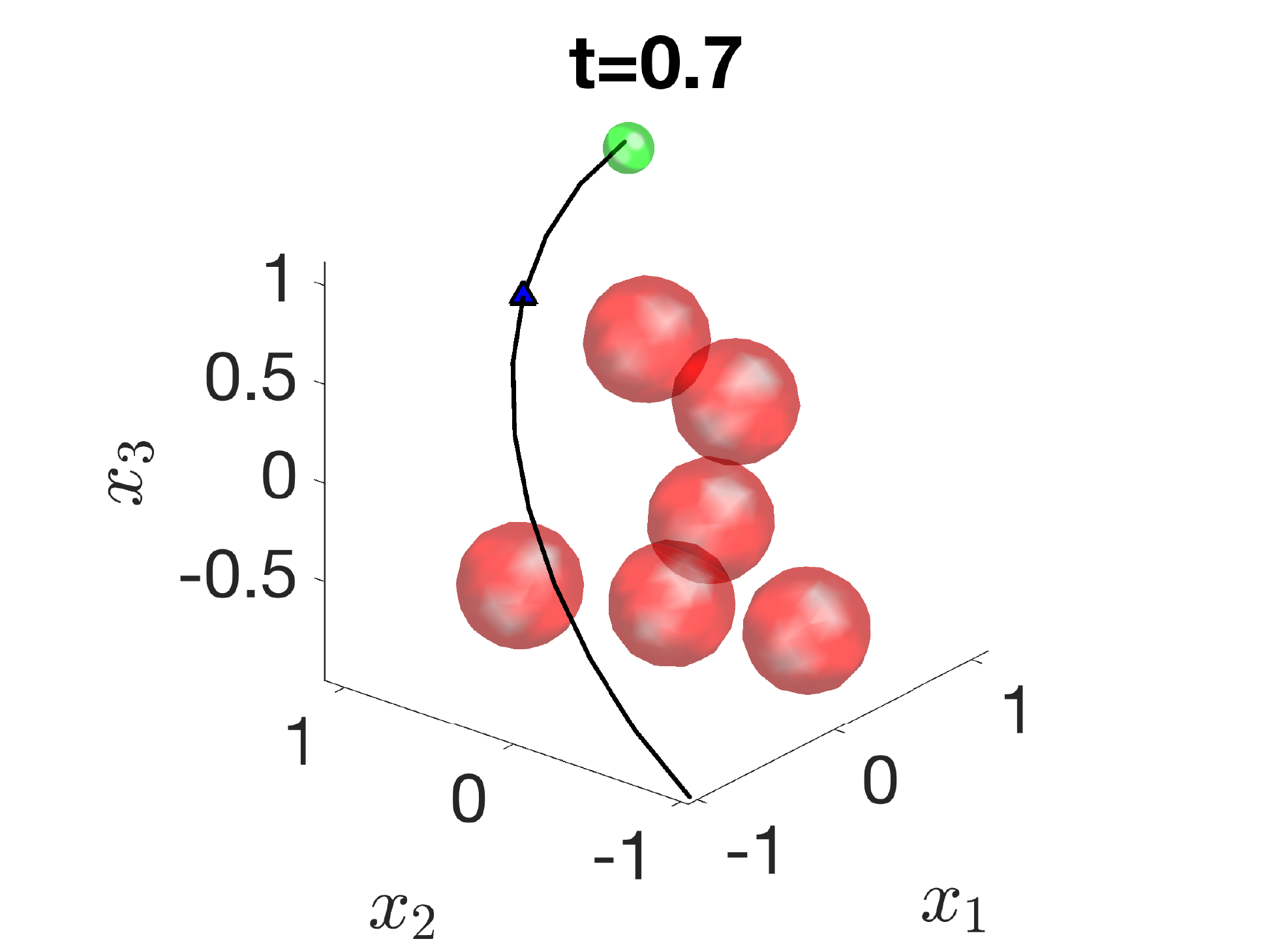}
    \includegraphics[width=0.2\textwidth]{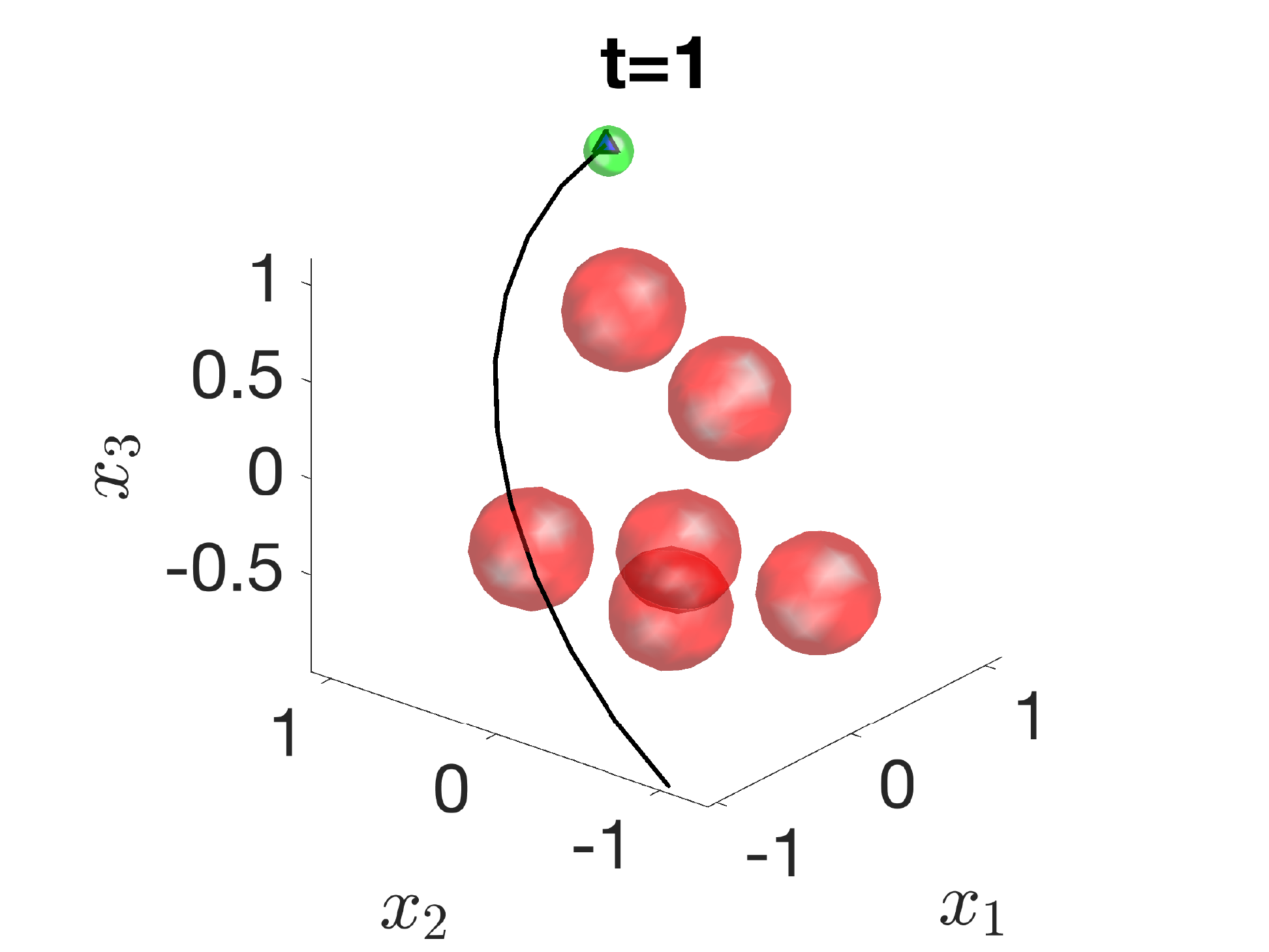}
    \caption{Example \ref{exp2}: Time steps $t=0.3, 0.4, 0.7$, and $1.0$. The black curve is the planned trajectory. The triangle is the expected system state. The red balls are moving obstacles. The green ball is the goal region. The trajectory has the bounded risk of 0.1 at each time step, and the system reaches the goal region with probability of at least 0.9.}
    \label{fig:exp2}
\end{figure}

The motion of an aerial vehicle in 3D space in the presence of wind disturbances is modeled as
\begin{align*}
    &x_{t+1} = x_{t} +\Delta T(v_t + {w_v}_t) \sin(\theta_t + {w_{\theta}}_t) \cos(\phi_t + {w_{\phi}}_t)\\
    &y_{t+1} = y_{t} + \Delta T (v_t + {w_v}_t) \sin(\theta_t + {w_{\theta}}_t)\sin(\phi_t + {w_{\phi}}_t)\\
    &z_{t+1} = z_{t} + \Delta T (v_t + {w_v}_t) \cos(\theta_t + {w_{\theta}}_t)
\end{align*}
where $(x_t, y_t, z_t)$ is the 3D position, ($v_t, \theta_t$, $\phi_t$) are the control inputs, and ($\omega_{v_t}, \omega_{\theta_t}$, $\omega_{\phi_t}$) are the external disturbances. The planning horizon is $T = 10$ and $\Delta T = 0.1$. The noises ${w_v}_t$, ${w_{\theta}}_t$, and ${w_{\phi}}_t$ have uniform distribution over $[-0.1,0.1]$. The initial states $x_0, y_0, z_0$ have uniform distribution over $[-1-0.05, -1+0.05]$. There are 6 moving 3D uncertain obstacles in the environment. We model obstacles as moving 3D balls with uncertain radius and position. The radius of each ball has a uniform distribution over $[0.2, 0.4]$ and the trajectory of each ball is disturbed by Gaussian noise of mean 0 and variance 0.001 in each axis. We want to steer the system to the goal region around $(1,1,1)$, respecting the risk bound $\Delta = 0.1$. 
The optimization has around 400 variables and around 600 constraints.
It took the solver around 7 seconds to find a solution. 
The obtained trajectory is plotted in \Cref{fig:exp2}.
We estimate the risk using Monte Carlo simulation. We sample one million points from the initial distribution and simulate them forward via the stochastic dynamics.
The trajectory is verified to have the bounded risk of 0.1 at each time step, and the system reaches the goal region with probability of at least 0.9.

\subsection{Ground Vehicles}\label{exp4}
\begin{figure}[t]
    \centering
    \includegraphics[width=0.23\textwidth]{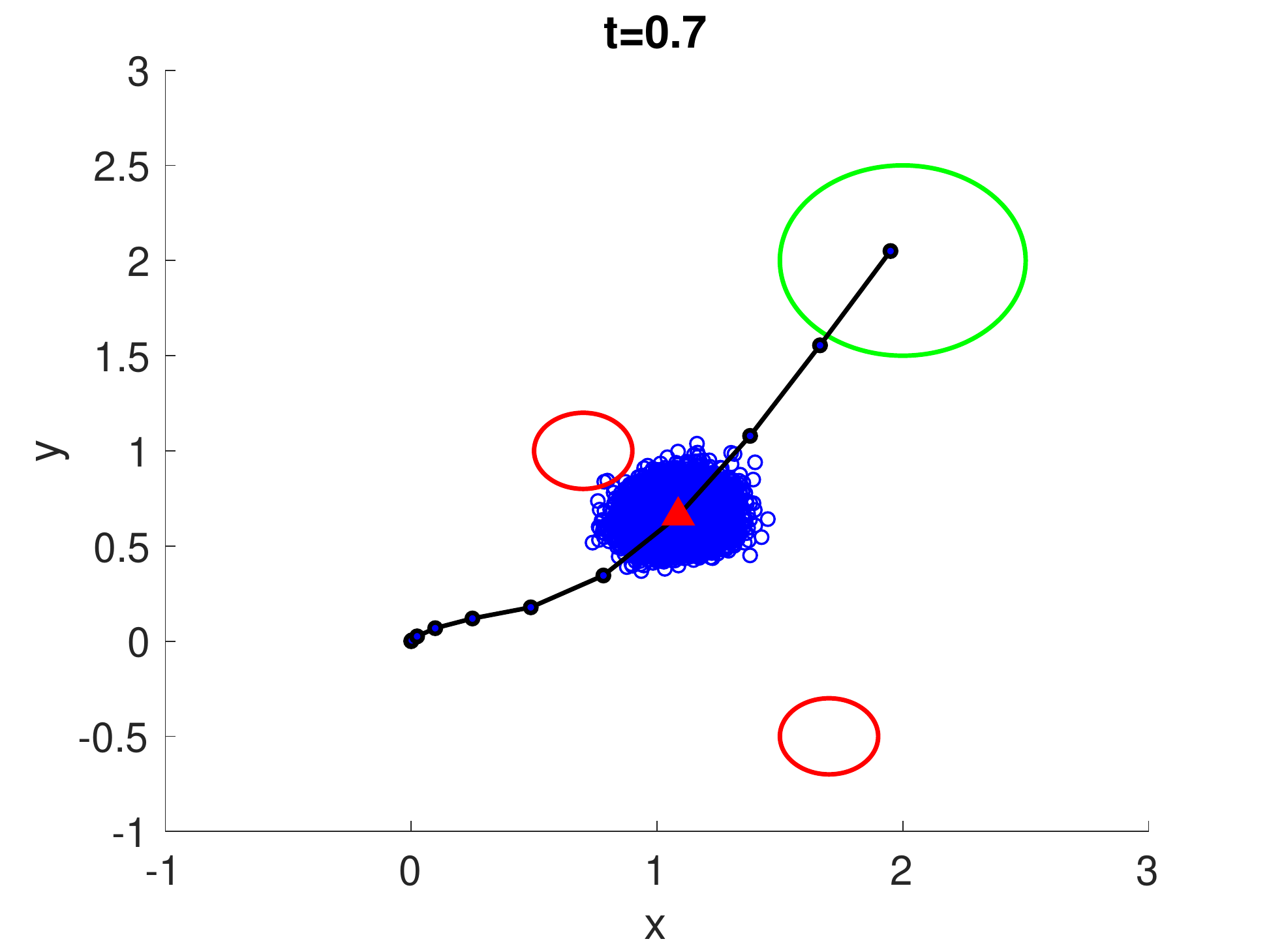}
    \includegraphics[width=0.23\textwidth]{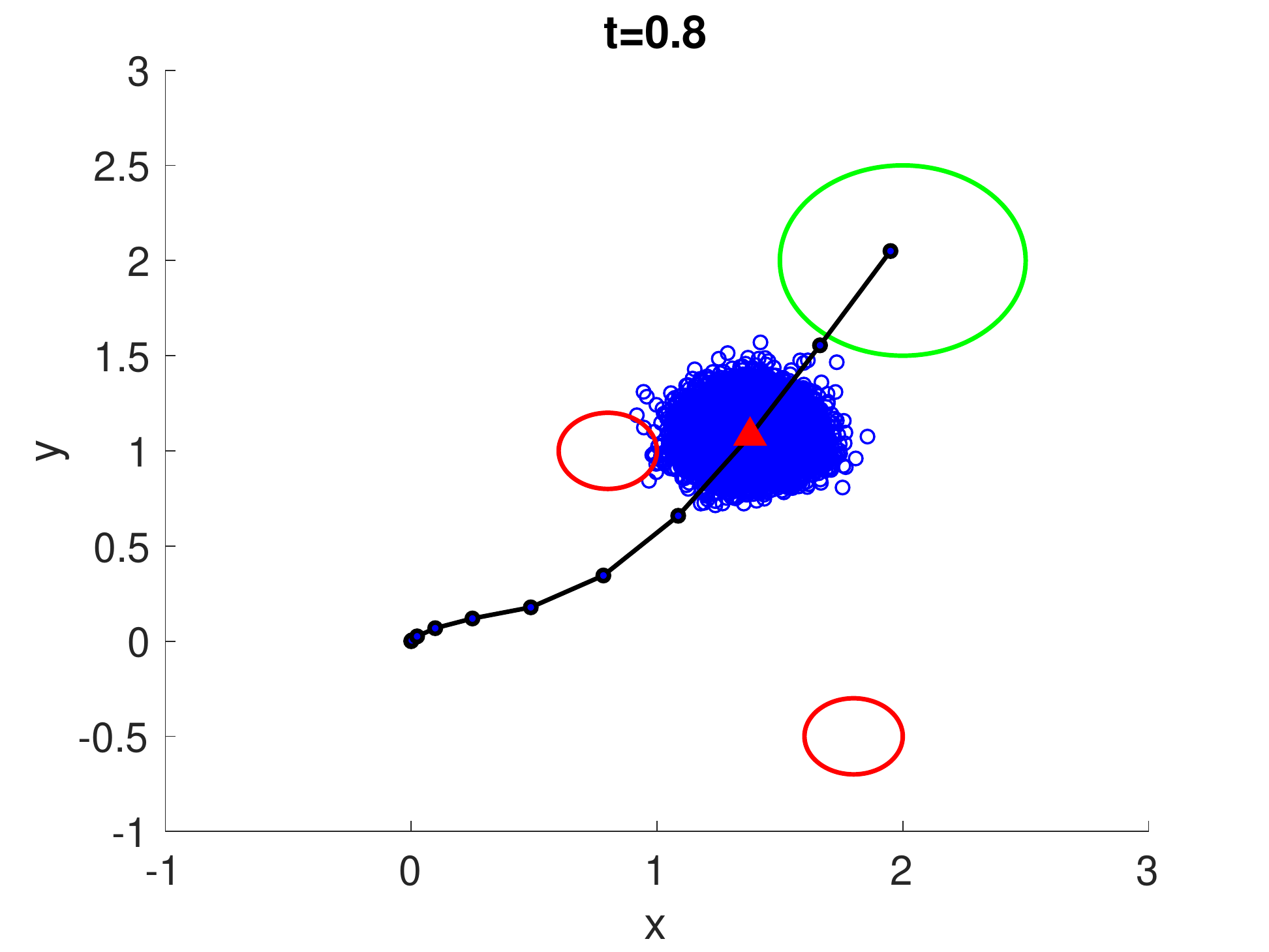}
    \includegraphics[width=0.23\textwidth]{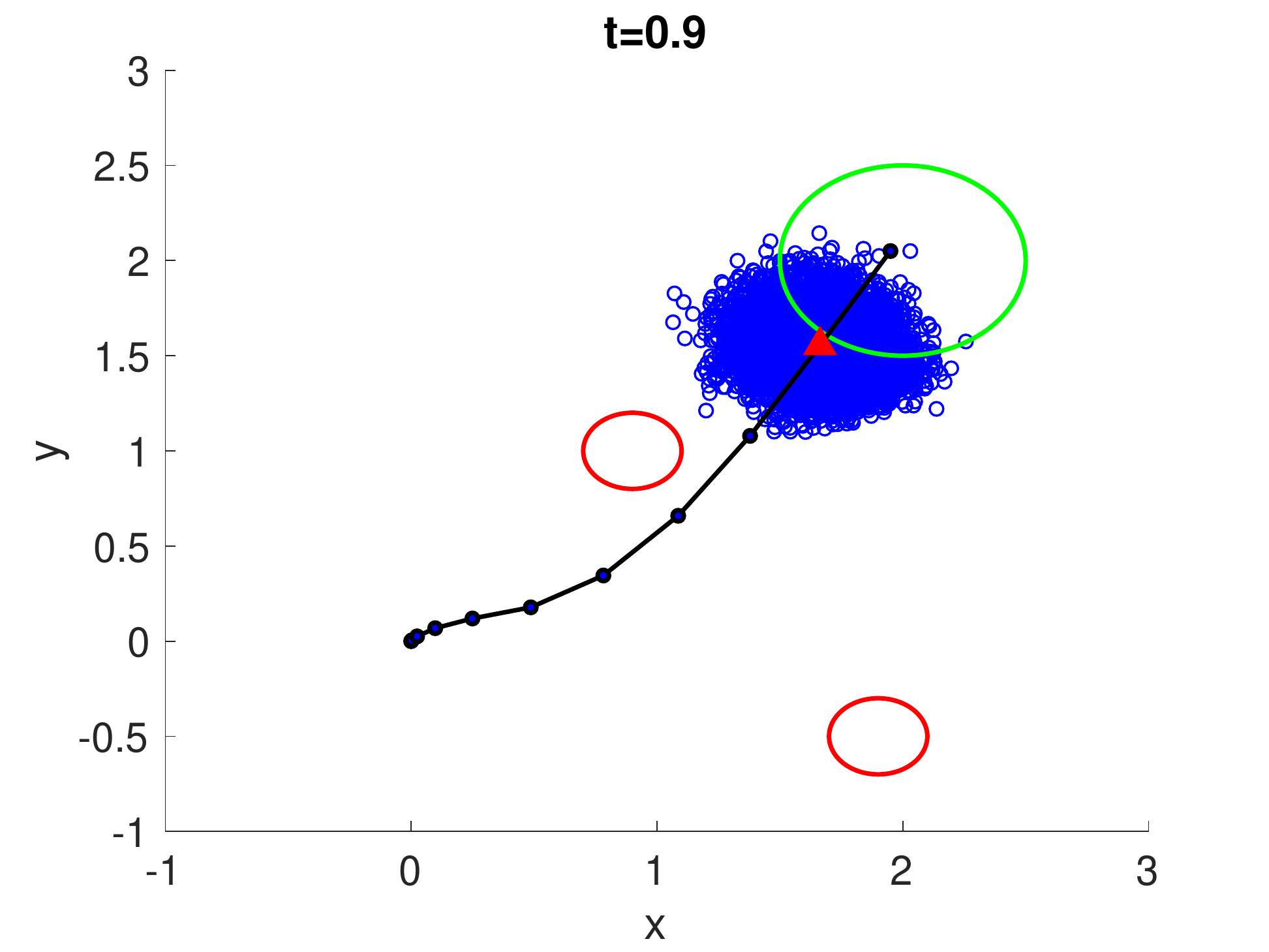}
    \includegraphics[width=0.23\textwidth]{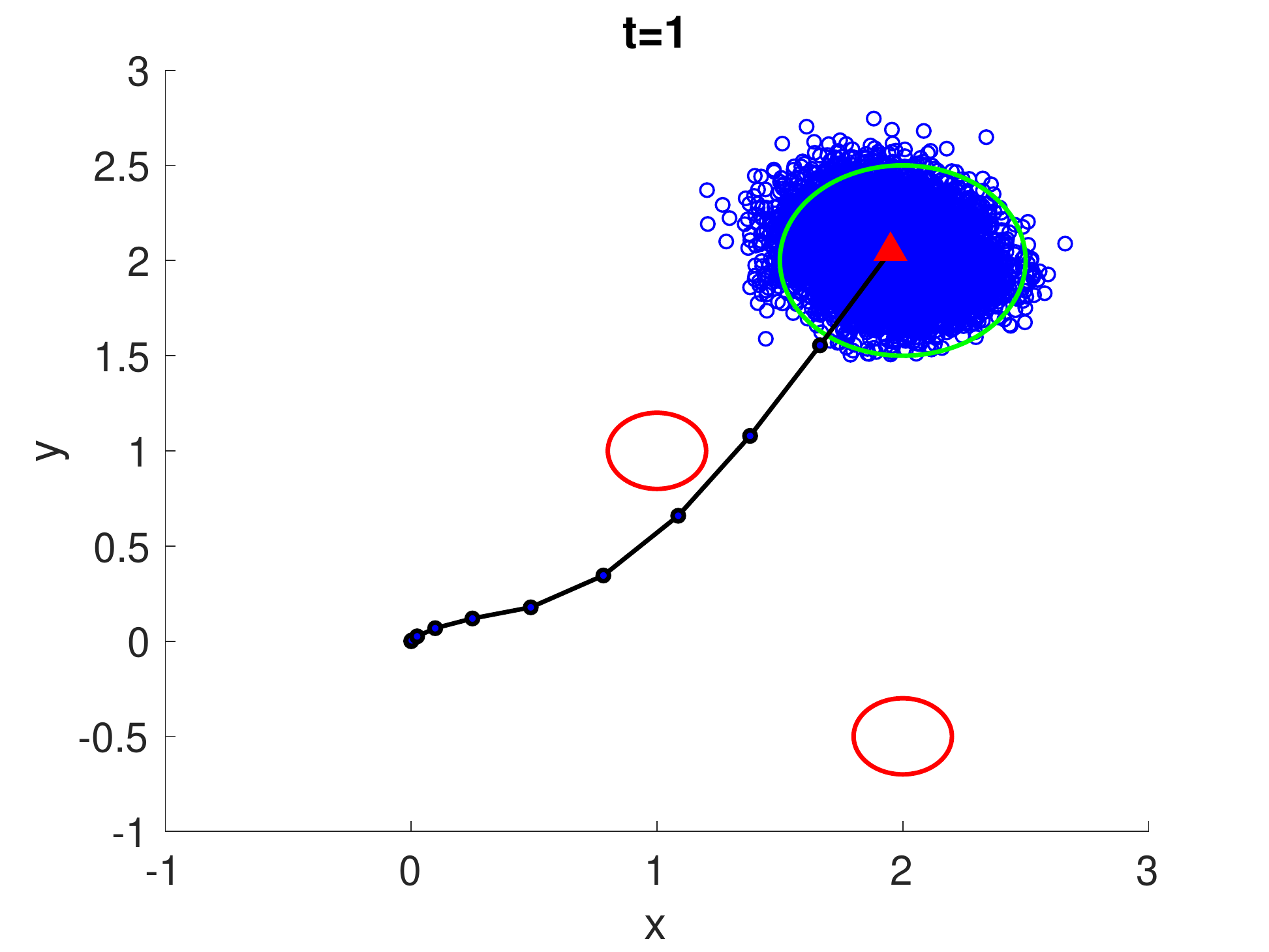}
    \caption{Example \ref{exp4}: Time steps 0.7, 0.8, 0.9, and 1. The black curve is the planned trajectory. The blue circles are 10,000 samples of system states. The red curves are moving obstacles. The green curve defines the goal region. The red triangle is the current expected state on the trajectory. The system has the bounded risk of 0.1 and enters the goal region with probability of at least 0.9.}
    \label{fig:exp4}
\end{figure}

In this example, we consider a ground vehicle model with the following uncertain dynamics 
\begin{align*}
    &x_{t+1} = x_{t} + \Delta T v_t  \cos(\theta_t)\\
    &y_{t+1} = y_t + \Delta T v_t \sin(\theta_t)\\
    &v_{t+1} = v_t + \Delta T (a_t + {\omega_{v_t}})\\
    &\theta_{t+1} = \theta_t + \Delta T (u_t + \omega_{\theta_t})
\end{align*}
The states are the 2D position $(x_{t}, y_t)$, the velocity $v_t$, and the orientation $\theta_t$. The control inputs are $a_t$ and $u_t$. 
The planning horizon is $T = 10$ and $\Delta T = 0.1$. The noise $\omega_v$ has normal distribution with mean 0 and variance 1, while $\omega_{\theta_t}$ has Beta distribution with parameters $(1,3)$ over $[0,1]$. The initial states $x_0$, $y_0$, $v_0$, and $\theta_0$ have uniform distributions over $[-0.05, 0.05], [-0.05, 0.05], [0, 0.1]$, and $[\pi/3-0.1, \pi/3+0.1]$, respectively. The environment has two moving vehicles with uncertain locations. One vehicle is defined by $p_1(\mathbf{x},t) = (x_1-(\omega_1+t))^2+(x_2-1)^2-0.2^2\leq 0$, where $\omega_1$ has uniform distribution over $[-0.1,0.1]$. The other vehicle is defined by $p_2(\mathbf{x},t) = (x_1-(1+\omega_2+t))^2+(x_2+0.5)^2 -0.2^2\leq 0$, where $w_2$ has uniform distribution over $[-0.05, 0.05]$. 
The optimization has around 2,300 variables and around 2,400 constraints. 
It took the solver around 5 minutes to compute the solution. 
The obtained trajectory is represented by the black curve in \Cref{fig:exp4}. To verify the results, we sample  one  million  points  from  the  initial distribution  and  simulate  them  forward  via  the  stochastic dynamics shown in \Cref{fig:exp4}.
The system is verified to have the bounded risk of 0.1 at each time step and enters the goal region with probability of at least 0.9. 

%% file: sec/conclusion.tex
We have proposed a method to solve the most general motion planning problem where the system dynamics is stochastic and nonlinear, the initial position is uncertain, the uncertain obstacles can move and change shapes over time, and all the uncertainties are not necessarily Gaussian. Our method builds on moment-based representation and propagation of uncertainties to convert the probabilistic trajectory planning problem into a deterministic nonlinear optimization where the off-the-shelf nonlinear solvers can be used to obtain the optimal solutions. 
For the future work, we will use a similar framework to design feedback controllers to track the nominal trajectory while reducing the tracking uncertainty of the system states over the planning horizon. 